# Event-based tracking of human hands

Laura Duarte, Mohammad Safeea, Pedro Neto

*Universidade de Coimbra*

**Abstract**

**Purpose** – Human hands tracking using data from event camera (dynamic vision sensor). The event camera detects changes in brightness, measuring motion, with low latency, no motion blur, low power consumption and high dynamic range. Captured frames are analysed using lightweight algorithms reporting hands position data to be used in human-robot interaction and robot safety applications.

**Design/methodology/approach** – Events data are pre-processed, noise is reduced, the regions of interest (ROI) are defined considering boundary events activity, and ROI features (hands tracking in space) are extracted.

**Findings** – Event-based tracking of human hands demonstrated feasible, in real-time and at a low computational cost. The proposed ROI-finding method reduces noise from intensity images, achieving up to 96% of data reduction in relation to the original, while preserving the features. The error in relation to ground truth (measured with wearables), measured using Dynamic Time Warping (DTW) and using a single event camera, is from 10 to 40 millimetres, depending on the plane it is measured.

**Originality** – Tracking of human hands in 3D space using a single event camera data and lightweight algorithms to define ROI features (hands tracking in space).

# 1. Introduction

## 1.1. Motivation and proposed approach

The scientific community has been putting a lot of effort into developing reliable computer vision systems through which robots perceive the real world surrounding them, including humans. Computer vision sensing provides information related to humans/objects pose and motion in space, including features that may conduct to the classification of such objects. This is critical for robot autonomy, human-robot interaction (HRI) and human-robot safety. As such, different vision sensors have been developed, many times supported by powerful machine learning methods featuring human/object classification and tracking.

Visual data can be acquired by frame-based cameras, structured-light 3D scanners, thermographic cameras, Laser Dynamic Range Imager (LDRI) devices, among others. Although these sensors have already been proven in many applications, they still present relatively high latencies, motion blur, high power consumption, low dynamic range, outputting large amounts of data (frames) which need to be transmitted, stored, and classified at a high computational cost. Each sensor has its own relative advantages and disadvantages (Yahya *et al.*, 2019). Some wearables present drift and are sensitive to magnetic disturbance while non-wearables like vision suffer from occlusions (Mokhtari *et al.*, 2018).

A new vision sensor overcomes some of the above issues, the Dynamic Vision Sensor (DVS), commonly named event camera (Lichtsteiner, Posch and Delbruck, 2008). The event camera detects changes in brightness, measuring motion, with low latency, no motion blur, low power consumption and high dynamic range. In such a context, captured events are analysed using lightweight algorithms featuring low computational cost, which pave the way for several new applications in robotics and other domains (Gallego *et al.*, 2019). The DVS has been applied for motion tracking, demonstrating adaptability and robustness to light-varying conditions and unstructured backgrounds. For example, impressive results have been reported in quadrotors featuring dynamic obstacle avoidance (Falanga, Kleber and Scaramuzza, 2020). Nevertheless, its capabilities are not yet fully explored, namely in what concerns to human tracking.

Aiming for human hand tracking, a main input for many robotic applications, in this study we propose to use video data from a single event camera as interface. Events data are pre-processed, noise is reduced, the regions of interest (ROI) are defined considering boundary events activity, and ROI features are extracted, Fig. 1. Events data are transformed into event frames with constant intervals of events which, by condensing data into relatively small data matrices, are much more manageable computationally than the initial captured real-time event stream. A novel algorithm is proposed to automatically define the hand ROI in data stream, resulting on intrinsic features which are used to track the hand in space. The hands trajectories are smoothed using moving average method. The proposed approach for hands tracking is evaluated by comparing it with other sensors (wearables), a magnetic tracker and a vision + inertial measurement unit (IMU) tracker.

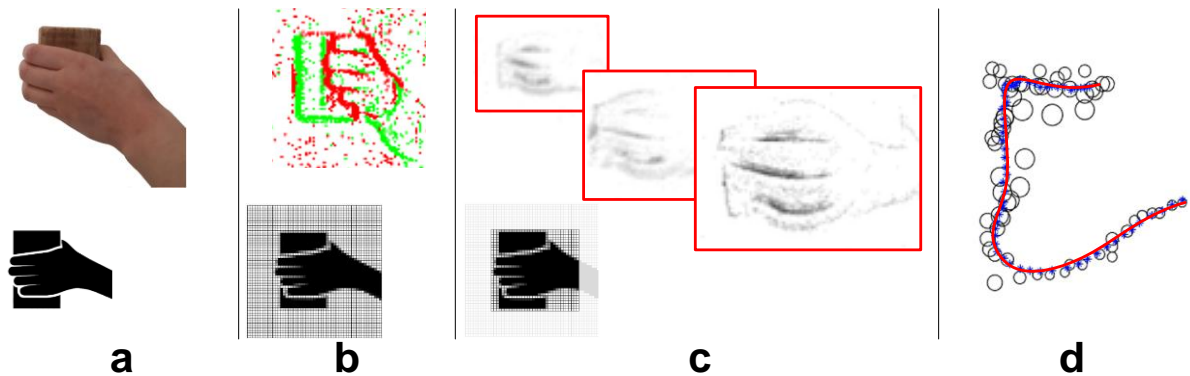

Figure 1: Overview of the proposed approach. a) Real-time event stream to capture hand motion when grasping an object. b) Intensity image representation of *n* events. c) Definition of Region of Interest (ROI) and feature extraction. d) Hand trajectory estimation from features and smoothing using the moving average.

### 1.2. Related studies

One of the first showcases of the DVS towards high-speed (low latency) applications was the use of two DVSs to control an actuated table to balance a pencil (Conradt *et al.*, 2009). This was achieved by continuously updating the estimate of the line representing the pencil using Hough Transform. As two DVSs are used, this allows for the estimation of the pencil's position in 3D space. Other DVS-based studies have emerged in the field of high-speed robotics. A representative example is proposed by (Mueggler, Huber and Scaramuzza, 2014), where the DVS sensor is used to successfully estimate the 6-DOF localization of a quadrotor during high-speed manoeuvres. It resorts to the Hough Transform and tracking changes to the line segments of a pattern. Another display of DVS's high-speed capability is the 3D particle tracking in a wind tunnel achieved in (Borer, Delbruck and Rösgen, 2017) with the use of three synchronized DVSs and Kalman filters.

In (Mitrokhin *et al.*, 2018) a method for producing a motion-compensated event stream was proposed to effectively distinguish events related to objects that move independently to the camera from events generated by the DVS's movement. The concepts of this algorithm were successfully used for quadrotor obstacle avoidance (Falanga, Kim and Scaramuzza, 2019). Feature tracking has also been studied using the estimation of velocity of non-specified features (Dardelet, Ieng and Benosman, 2018) and contour motion (Barranco, Fermuller and Aloimonos, 2014).

A popular approach to event data clustering for object tracking has been the use of concepts related to the mean shift algorithm (Fukunaga and Hostetler, 1975). One of its first usages on event data, (Litzenberger *et al.*, 2006), adapts the algorithm to create circular event regions with a timestamp resolution of 1 ms, which enables tracking of vehicles and people. In a more recent attempt, (Barranco, Fermuller and Ros, 2018), does not accumulate events into event frames, taking better advantage of the DVS high temporal resolution while showing great tracking results. Other Expectation-Maximization (EM) algorithms were used in (Zhu, Atanasov and Daniilidis, 2017) and in (Piątkowska *et al.*, 2012), where feature tracking in DVSs is achieved through probabilistic data association.

A step towards the DVS application in human-machine interfaces is the real-time gesture recognition system from event data (Amir *et al.*, 2017). By using an event-based processor, based on a Convolutional Neural Network (CNN), it manages to achieve a high recognition accuracy. Instead of directly using the continuous stream of events like the previous approaches, it collects sequences of events using a temporal filter cascade. In (Chen, 2018) a CNN was also used with fixed intervals of time for object detection, with the addition of a traditional camera synchronized with the DVS to generate pseudo-labels to train the CNN on DVS data. A capture system using both a DVS and a frame-based camera is also presented in (Saner *et al.*, 2014). The novel use of event coherence according to the event distribution at each moment is presented in (Wu *et al.*, 2019), processing multiple events at a time by creating "stream blocks".

In summary, in the last decade we witnessed an increasing number of studies with DVS systems. While DVS-based solutions are promising with clear advantages in relation to other sensors, its application is still limited. More research is needed, namely on the capture of features from events data which can be used for human tracking, robot navigation and human-robot interaction. In addition, the expected advantages of fusing events data with data from other sensors is not yet fully explored.

## 2. Data treatment

### 2.1. DVS data pre-processing

Each DVS event $e_i = (x_i, y_i, t_i, pol_i)$ contains information about its pixel coordinates, $x$ and $y$, the timestamp of the event, $t$, and the sign of the occurred brightness change, also known as polarity, $pol$. Additionally, events are asynchronously generated at a very high rate, making it a challenge to process them. As such, novel methods to specifically handle DVS data should be considered.

A possible solution relies on using an initial image which is continuously updated with events arriving. Its implementation requires dedicated techniques like the Hough Transform, but on the other hand this method can deal with the DVS's extremely low latency which must be harnessed to track high-speed movements. This approach is generally used for high-speed robotics applications, like quadrotors orientation tracking. Although this method provides low latency, it is difficult to use for motion tracking and object recognition, due to the inherent complexity of most object's shapes.

Other methods are based on the accumulation of events over a fixed interval of time into a frame, called an integrated image. Accumulating events is a solution fit for tracking movements that contain complex shapes, e.g. cars or hands. It has the additional advantage of allowing the adoption of already developed frame-based tracking and recognition algorithms. The discretization of the DVS continuous output into integrated images, although limiting the sensor's low latency and high temporal resolution, makes it easier to process data. Usually, integrated images are built using all the events generated in a predefined interval of time, Fig. 2 (a). However, for very slow movements a lot of redundant frames are captured and, on the other hand, for fast movements the framerate is limited. An update to this approach relies on the capture of a frame at each predetermined number of events, Fig. 2 (b). This approach is more versatile as it adapts the framerate to the speed of the movement, accounting for both fast and slow movements. For a high number of events per frame, the image generated at each frame is more recognizable but, due to the increase of data, it has more noticeable noise and lower framerate.

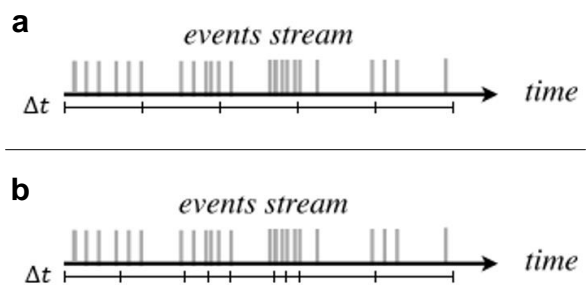

Figure 2: Integrated image build methods. a) Constant intervals of time. b) Constant intervals of events.

## 2.2. Noise Reduction

As in other sensors, DVS data need to be processed to reduce/remove as much irrelevant data (noise) as possible. It is challenging to differentiate DVS noise from relevant data, which tends to cause weak tracking solutions.

As DVSs capture high quantities of data, naturally, many of generated events have noise. For instance, any movement within the DVS range of view will be detected, even if such movement is unintentional or relatively small. This is visible when, for example, we are focused on the movement of the human hand but in the background the chest and the rest of the body also move, Fig. 3 (a), the moving hand/object casts a shadow, Fig. 3 (b), or there is no movement in the captured scene causing an accumulation of irrelevant events, Fig. 3 (c).

An approach to promote noise reduction is to ignore the events outside the image's region of interest (ROI). This type of image manipulation process, often called cropping, eliminates significant part of DVS noise.

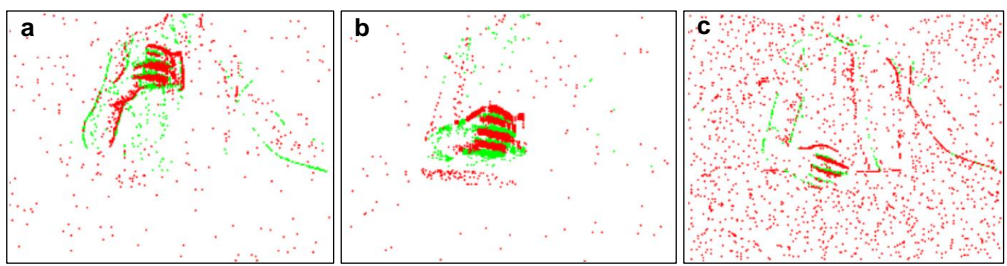

Figure 3: DVS noise sources where the green and red points represent events with positive and negative polarity, respectively. a) Body/object movement in background. b) Shadows. c) Idle.

## 3. Region of Interest (ROI)

### 3.1. Image creation from DVS data

We aim is to define a methodology to find the ROI within captured DVS data without any previous knowledge of the scene (object type or the proximity of the object to the DVS). Such ability is key for human tracking in the presence of unexpected changes in the environment.

Let us assume that at each new *n* events, an integrated image is created from the DVS video. The event data available for each image is represented by a matrix with dimensions 4 x *n* (4-dimensional event data). Rather than using the data in this format to find the ROI, it is instead converted into an image with the dimensions of the DVS's resolution, 240 x 180 pixels. This is accomplished by using Algorithm 1, which stores the number of times an event occurs at each possible coordinate into a matrix. This creates what is called an intensity image, containing values that range between the lowest and highest event counts within the matrix. As events are used to create the image, it can also be referred to as an event frame. In this study, both negative and positive polarity events are added to the count of each coordinate of the image, differing from previous works in which the events with negative polarity subtract instead. Fig. 4 shows how each integrated image is constituted. Changing the DVS data to an event frame has the advantage that the regions of the image with higher event counts can be defined as the ROI. This is true since the movement intended to be captured by the DVS tends to generate a lot more events than noise.

---

**Algorithm 1:** Creation of an Intensity Image at each $n$ events

---

**Input:** Frame of events $e_i = (x_i, y_i, t_i, pol_i)$, Number of events per frame $n$
**Output:** 240 x 180 pixel intensity image $I(x, y)$

1   Zero initialization of $I(x, y)$ with dimensions 240 x 180 and $counter := 0$
2   **for each** $e_i$ **do**
3      $I(x_i, y_i) := I(x_i, y_i) + 1$
4      $counter := counter + 1$
5      **if** $counter = n$ **then**
6         Return $I(x, y)$

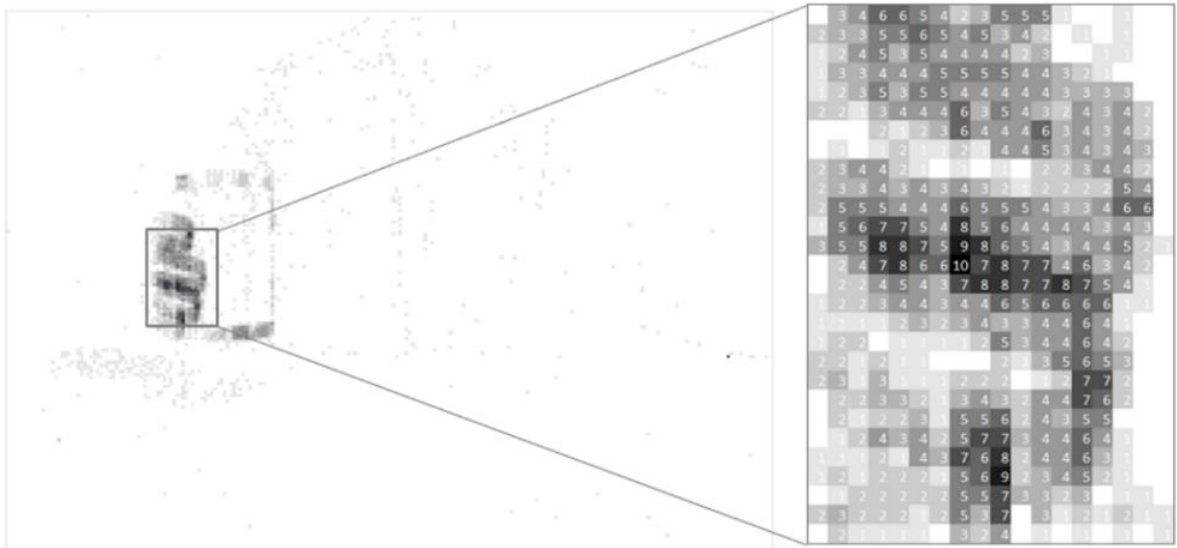

Figure 4: Representation of an integrated image created from 3000 events, with a magnified section.

### 3.2. Events per Frame

It is proposed a novel methodology to define the number of events per frame that better serves our objective (ROI tracking). It is based on the analysis of a given tracking sequence (pick and place of an object) performed at both slow and fast motion and considering different number of events per frame. We defined a quality classification method where according to the image quality a score from 1 to 5 is given to each criterion (1 is the worst and 5 is the best). The considered criterions are the recognition of the hands, the absence of random noise, the absence of movement related noise and the number of frames per second. The obtained scores resulted in a weighted decision matrix, Table I. From Table I, it can be concluded that for 3000 events per frame we have the highest score and, as such, it will be the value used in this study. Fig. 5 shows captured frames obtained with 3000 events per frame.

Table I: Weighted decision matrix to define the number of events per frame.

| Criterion | Weighting | Events per frame | | | | | | | |
|---|---|---|---|---|---|---|---|---|---|
| | | 1000 | | 2000 | | 3000 | | 4000 | |
| | | Slow | Fast | Slow | Fast | Slow | Fast | Slow | Fast |
| Recognition of hands | 5 | 2 | 2 | 4 | 3 | 5 | 4 | 5 | 4 |
| Absence of random noise | 4 | 2 | 1 | 3 | 2 | 4 | 3 | 4 | 3 |
| Absence of movement related noise | 3 | 5 | 5 | 4 | 4 | 3 | 3 | 2 | 3 |
| Frames per second | 2 | 5 | 5 | 3 | 3 | 3 | 3 | 2 | 2 |
| TOTAL | | 82 | | 91 | | 103 | | 96 | |

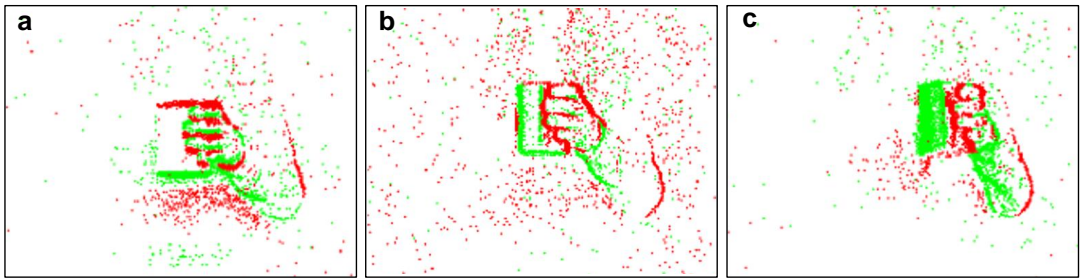

Figure 5: Captured frames with 3000 events per frame in a pick and place task of a cylinder-shaped object. The green and red points represent events with positive and negative polarity, respectively.

### 3.3. Creation of the ROI

The DVS camara perceives the environment in 2D as a closed shape, limited by its boundaries. The edges of the object in relative motion to the camera are highlighted as the brightness change is much more significant than within the object's shape, Fig. 6. From the DVS intensity image, it can be derived the object's contour defining the ROI.

Many objects have vertically aligned boundaries, and this is the reason why a rectangular format is typically used as the shape of the ROI. As such, locating the outlines in intensity images is accomplished by finding columns of pixels with high intensity values. The same is true for finding horizontally aligned outlines by using rows instead. A straightforward method of finding these high-count columns and rows relies on the sum of the pixels' intensity values for each column and row of the image. In this scenario, we propose an effective method to define the ROI limits according to the previously obtained sums of the pixels' intensity. For the sake of clarity, the process to estimate the vertical limits of the ROI is here detailed, and a similar approach applies to the estimation of the horizontal limits.

Having the sums of each column, one could consider taking the average of these sums and finding the first and last instances in which the sum is greater than the average value, effectively using it as a threshold, Algorithm 2. The result/output of this algorithm on an event frame is represented in Fig. 7. It shows reasonably good results in the creation of the ROI, considering the simplicity of the algorithm. Only unnecessary information is eliminated, and the image area is reduced by 77%. The proposed approach demonstrated to be able to adapt to event frames created from different kinds of objects and environments.

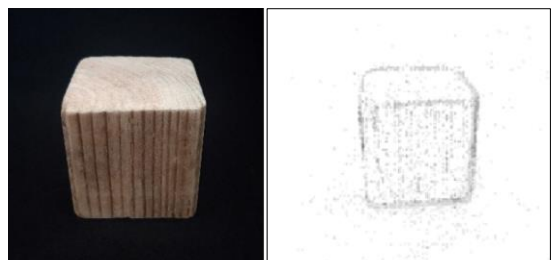

Figure 6: Wooden cube (left) and event camera integrated frame captured with 3000 events per frame (right).

**Algorithm 2:** Creation of the ROI: First and last instance values greater than the average

**Input:** 240 x 180 pixel intensity image with 3000 events $I(x, y)$
**Output:** Values of left and right vertical ROI boundaries, $x_{min}$ and $x_{max}$

1. Obtain the average of the image's column intensity values $\mu_{\sum I(x)}$
2. **for each** $x$ **do**
3.    **if** $\sum I(x) > \mu_{\sum I(x)}$ **then**
4.       **if** $x_{min} = 0$ **then**
5.          $x_{min} := x$
6.       **else if** $x > x_{max}$ **then**
7.          $x_{max} := x$

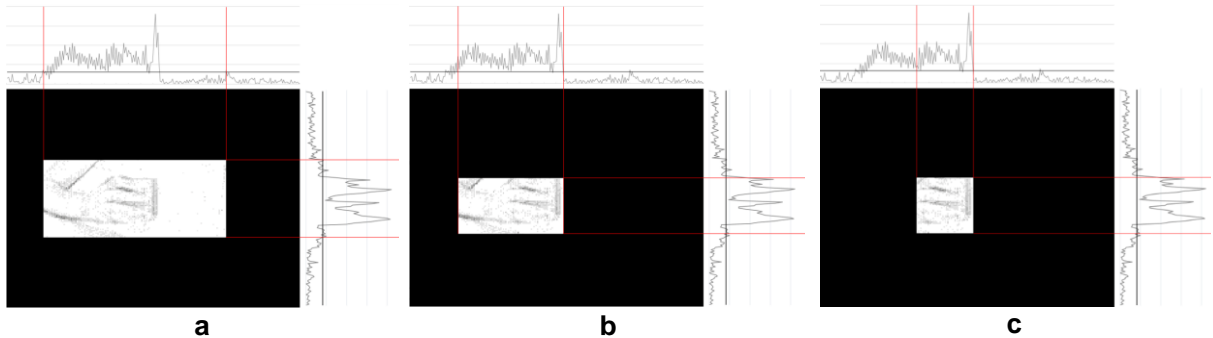

Figure 7: a) ROI boundary obtained from Algorithm 2. b) ROI boundary applying the first rule to Algorithm 2. c) ROI boundary obtained from Algorithm 3.

Even in the presence of promising results using the above method (Algorithm 2), it still struggles with instances of columns that slightly surpass the average due to noise. In such a context, two additional rules were added to previous method. A first rule is implemented to prevent noise spikes from triggering the start or end of the ROI, considering that at least *c* consecutive column sums with values greater than the average must exist to start and end the ROI. A value of *c*=3, Fig. 7 (b), reduces the image area by 90% while preserving the hand/object on the ROI. The second rule ensures that the resulting ROI is square, helping with a further reduction of non-essential information, achieving 96% data reduction of the original intensity image. Fig. 7 (c) shows the resulting ROI by applying Algorithm 3 (both rules implemented).

The proposed methodology to automatically create the ROI manages to successfully differentiate the hand/object from noise. Additionally, this methodology only needs slight adaptations to enable the detection of multiple ROI zones, for example both human hands in motion. Within a mostly static scene, two different behaviours for the algorithm have been observed. When the image has no distinct ROI due to sparse events, the algorithm slightly crops the image, or does not crop the image at all. This behaviour can be used to identify static environments.

Another important issue of the proposed algorithm is related to the distance of the DVS to the human/object. The closer the human/object is to the DVS, the more difficult it is to detect its outline, as it is defined by more sporadic events.

**Algorithm 3:** Creation of the ROI: First and last $c$ consecutive values greater than the average

**Input:** 240 x 180 pixel intensity image with 3000 events $I(x, y)$, Number of consecutive values above average required $c$
**Output:** Values of left and right vertical ROI boundaries, $x_{min}$ and $x_{max}$

1. Obtain the average of the image's column intensity values $\mu_{\sum I(x)}$
2. **for each** $x$ **do**
3.     **if** $\sum I(x) > \mu_{\sum I(x)}$ **then**
4.         **if** *all c following column sums* $> \mu_{\sum I(x)}$ **then**
5.             **if** $x_{min} = 0$ **then**
6.                 $x_{min} := x$
7.             **else if** $x > x_{max}$ **then**
8.                 $x_{max} := x$

9. $x_{range} := x_{max} - x_{min}$
10. **if** $x_{range} > y_{range}$ **then**
11.     **for** $x_{min} \leq x \leq x_{max} - y_{range}$ **do**
12.         Save $x$ corresponding to biggest value of $\sum_{n=x}^{x+y_{range}} \sum I(n)$ as new $x_{min}$
13.     $x_{max} := x_{min} + y_{range}$

### 3.4. ROI features

From the image ROI, we can extract features, namely its centre coordinates, Fig. 8, and the ROI size. When the human hand is stopped (no hand motion), the ROI is composed by a reduced number of events and, as such, the number of active pixels within the ROI is also considered as a feature. This feature validates if the ROI is capturing the human hand motion or another unrelated motion.

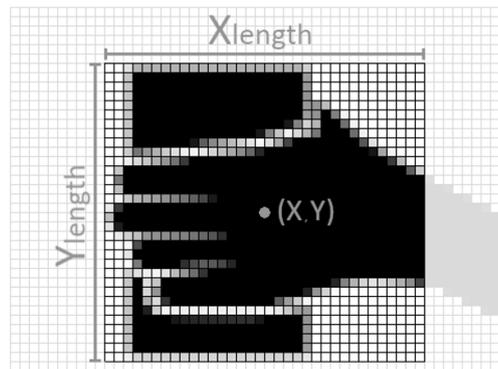

Figure 8: ROI features: centre coordinates and size.

## 3.5. Trajectory smoothing

The above features are used to estimate the hands trajectories, *x*, *y*, and *z* coordinates obtained for each event frame. The *x* and *y* coordinates are obtained from the ROI centre coordinates, while the *z* coordinates are obtained from the depth estimation of the ROI. This is achieved by analysing the size of the ROI, i.e., the closer the hand is to the DVS, the bigger the ROI size. Implementing proper calibration, we can estimate the *z* coordinate from the ROI size. Moving average is applied to the resulting coordinates to smooth the trajectory, making it more stable and countering sporadic irregular behaviours. Focusing solely on *x* coordinates we have:

$$x_i = \frac{1}{l}\sum_{j=1}^{l} ROI_{centre, x_{i-j}} \tag{1}$$

Where $x_i$ is the estimated value of the *x* coordinate at index position *i*, $l$ is the length of the sequence, and $ROI_{i-j}$ is the ROI's *x* centre coordinate at index position *i-j*. A value of $l = 20$ was used.

To avoid using non-relevant event frames, only the data of event frames that have more than a certain number of active pixels within the ROI are used. A minimum value of 20% of the pixels should be active within the ROI to count as a valid event frame.

## 4. Experiments and results

### 4.1. Experimental setup

The experimental setup was built to test the ability of the proposed event-based tracking methodology to estimate the trajectory of the human's hand when moving an object (wooden cube).

The camera used to capture the event video footage is a DAVIS 240C, Table II. It is a 240x180 pixel DVS camera with simultaneous active pixel frame output. However, only events data were captured. The DVS captures the human upper body to accommodate for the user's range of motion while focussing as much as possible on the hands and object, Fig. 9. The distance between the camera and the user is 60 centimetres. The system performance is compared with the one obtained from wearable sensors which results are considered ground truth, a magnetic tracking sensor Polhemus Liberty and a vision + IMU system Intel RealSense Tracking Camera T265. The human hand holds the wooden cube and moves it with an up-down-front-back-right-left movement sequence, corresponding to Y-Z-X directions, respectively.

Table II: DAVIS 240C specifications.

| Resolution | 240 x 180 pixels |
|---|---|
| Dynamic Range | 120 dB |
| Minimum latency | ~ 12 us @ 1klux with optimized biases |
| Bandwidth | 12 MEvents/ second |
| Power Consumption | 5-14 mW (activity dependent) |

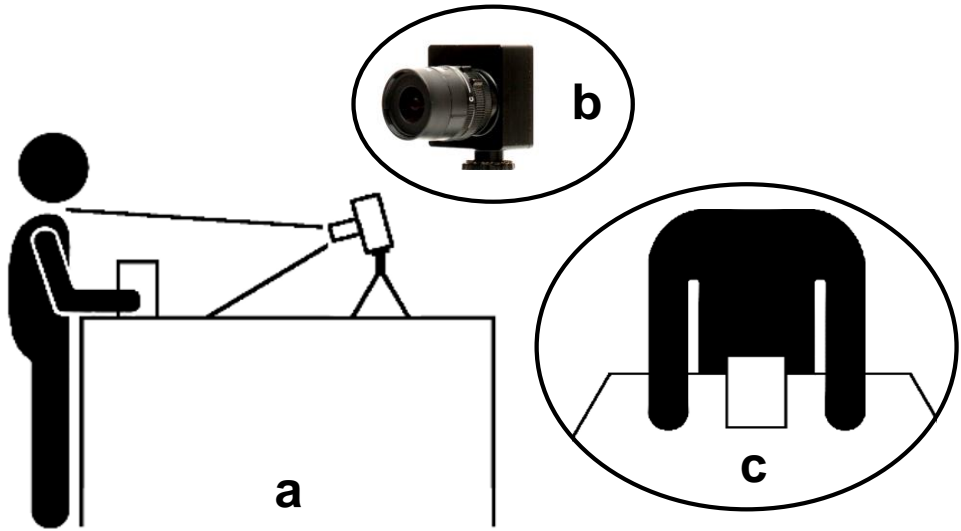

Figure 9: a) Recording setup. b) Real DAVIS 240C camera. c) DVS view.

## 4.2. Calibration

To find the correlation between the tracking coordinates obtained from the proposed methodology and the real-world coordinates, a video was recorded of a cube moving directly towards the camera while centred with the camera's optical axis, Fig. 10. By knowing the real and pixel sizes of the cube, we can estimate pixel to centimetres ratios for different distances to the camera and the approximated correlation between the ROI's size and the real distance of the object to the camera.

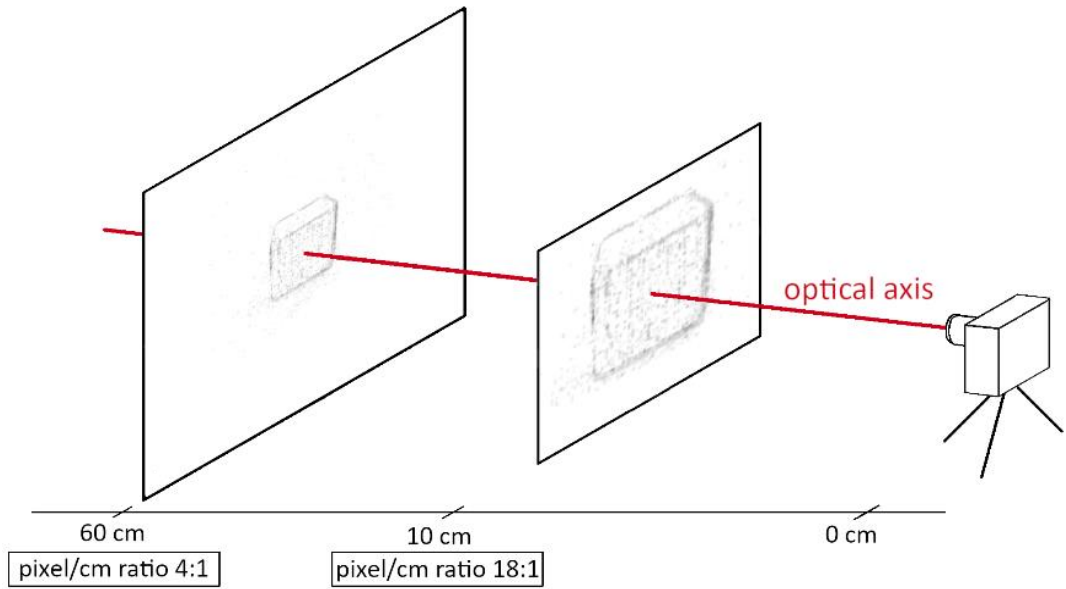

Figure 10: Pixel to centimetres ratios obtained for different values of distances between the object and the camera.

### 4.3. Results

Three different tests were conducted to evaluate the tracking accuracy of the proposed DVS-based tracking solution. Test 1 compares results obtained from the DVS and the magnetic tracking system, the latter of which is a wearable that will be treated as ground truth. Test 2 compares results obtained from the DVS and the vision + IMU system, which is another wearable that will also be handled as ground truth. Test 3 compares results obtained from the DVS and the magnetic tracking system (ground truth) in a different tracking motion focused on X-Y directions.

Wearables present disadvantages in relation to vision-based systems, such as the DVS, since they must be attached to the human body. Nevertheless, they tend to be more accurate. Hence, for comparison purposes, their outputs are considered the ground truth. The magnetic tracker presents an error of less than one millimetre in static conditions and controlled magnetic fields, while the vision + IMU has an error of 2%.

Test 1 shows that the proposed DVS-based system performs relatively well when the motion is performed perpendicular to the DVS's optical axis (X-Y plane), Fig. 11 (left). However, it struggles to estimate depth (along Z axis) through the size of the ROI, as its value fluctuates, even when the hand is at the same camera depth, Fig. 11 (right). Similar reasoning can be established from the results of test 2, Fig. 12. In test 3, the positional variation along Z axis is smaller, conducting to less error, Fig. 13. In this test the error in X-Y plane is relatively low.

From analysing the plots in Fig. 11, Fig. 12 and Fig. 13 it can be visually concluded that the error is significantly smaller when measured in X-Y plane. To quantify such error, we compared the points from the matching trajectories. Since they have different number of points, Dynamic Time Warping (DTW) is used to match points of the two trajectories in each test and estimate the error, expressed as the average per-point distances between the points of the trajectories, Table III. In the X-Y plane the error is about 10 millimetres while in X-Y-Z space the error varies from 25 to 40 millimetres. This is an indication that further research is required to better estimate depth tracking positions. Such error could be drastically reduced by using two DVS cameras.

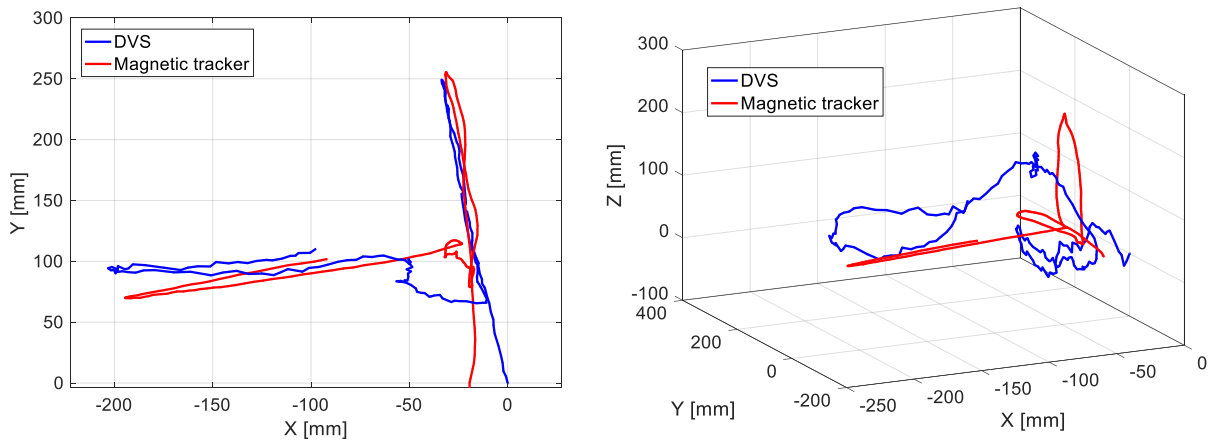

Figure 11: Test 1 results and comparison between trajectory obtained from DVS system and magnetic tracker.

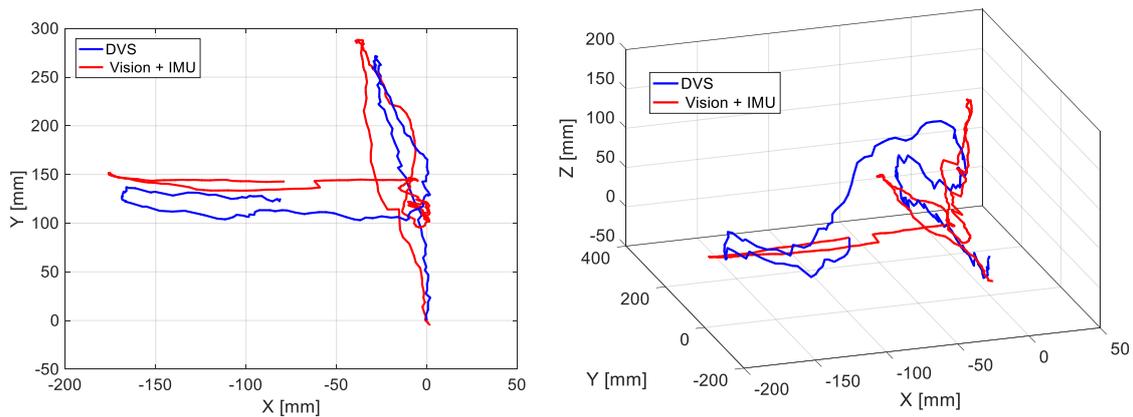

Figure 12: Test 2 results and comparison between trajectory obtained from DVS system and vision + IMU.

Table III: Trajectory error estimation expressed as average per-point distances.

|        | X-Y Error [mm] | X-Y-Z Error [mm] |
|--------|----------------|------------------|
| Test 1 | 14,1           | 37,1             |
| Test 2 | 12,97          | 43,4             |
| Test 3 | 11,74          | 24,9             |

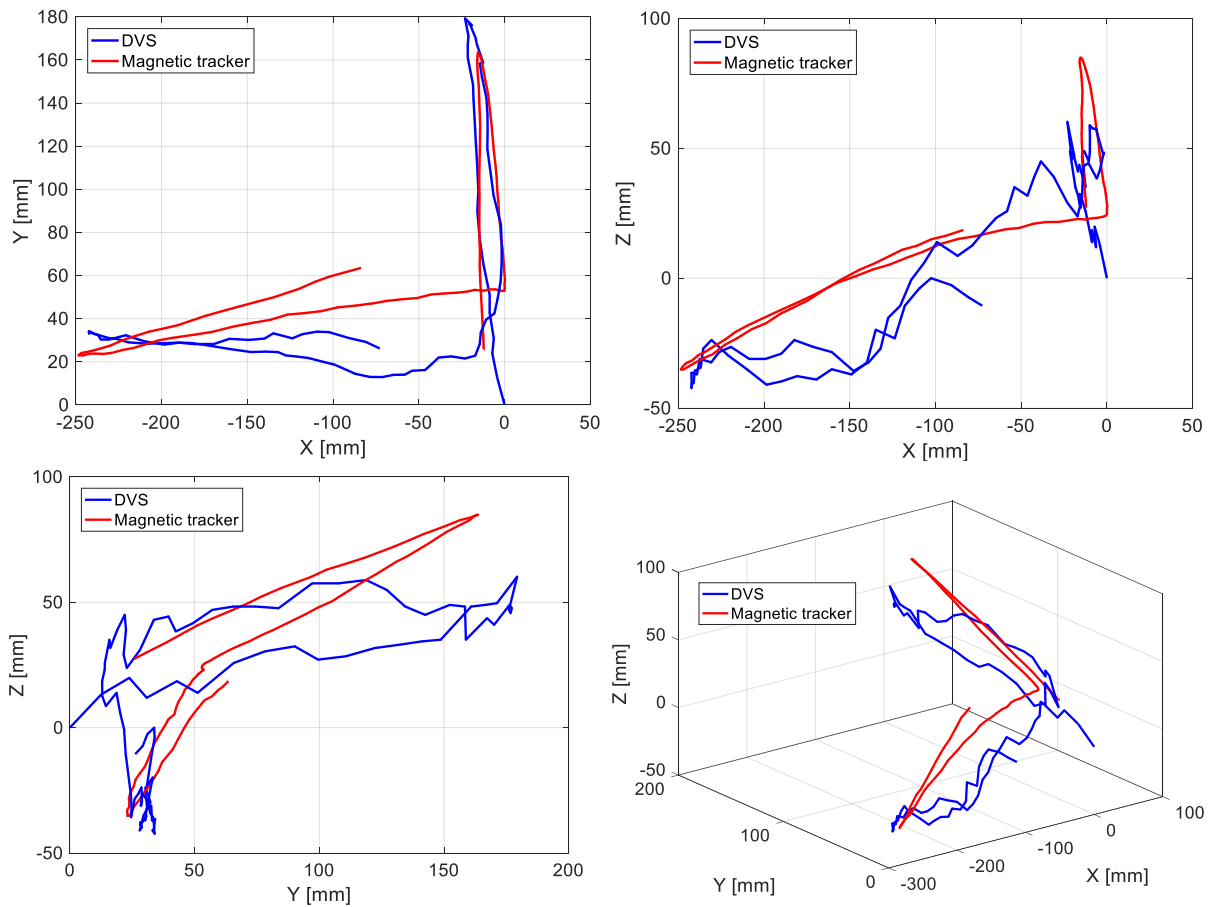

Figure 13: Test 3 results and comparison between trajectory obtained from DVS system and magnetic tracker.

## 5. Conclusion and future work

This study proposes a novel lightweight event-based method for human hands tracking using a single DVS camera. Events data are pre-processed, noise is reduced, the ROIs are defined considering boundary events activity, and ROI features (hands tracking in space) are extracted. The use of an event frame with constant intervals of events, instead of constant intervals of time, demonstrated better adaptation to both slow and fast movements. The proposed ROI-finding method removed noise from intensity images, achieving a maximum of 96% of data reduction in relation to the original, while preserving the image features. The extracted ROI's features demonstrated to be simple but effective towards the estimation of the localization of human's hands, mainly in the 2D space (X-Y plane) perpendicular to the DVS's optical axis. The error in X-Y plane is about 10 millimetres, while in X-Y-Z space the error varies from 25 to 40 millimetres

Further research is required to reduce the tracking error, mainly in the depth position estimation (along Z axis). Better results could be attained by using the knowledge of previous depth estimations to attempt to minimize large depth fluctuations and using supervised machine learning techniques to better estimate the ROI size and position.